\documentclass[letterpaper]{article} 
\usepackage{aaai23}  
\usepackage{times}  
\usepackage{helvet}  
\usepackage{courier}  
\usepackage[hyphens]{url}  
\usepackage{graphicx} 
\usepackage{natbib}  
\usepackage{caption} 
\usepackage{algorithm}
\usepackage{algorithmic}
\usepackage{amsmath,amssymb,amsfonts,eucal}
\usepackage{relsize}
\usepackage{subfigure}

\usepackage{newfloat}
\usepackage{listings}
\DeclareCaptionStyle{ruled}{labelfont=normalfont,labelsep=colon,strut=off} 
\floatstyle{ruled}
\newfloat{listing}{tb}{lst}{}
\floatname{listing}{Listing}
\nocopyright

\title{Representative Teacher Keys for Knowledge Distillation Model Compression Based on Attention Mechanism for Image Classification}

\author{
Jun-Teng Yang,\textsuperscript{\rm 1}
Sheng-Che Kao,\textsuperscript{\rm 1}
Scott C.-H. Huang,\textsuperscript{\rm 2}
}
\affiliations{
\textsuperscript{\rm 1} Institute of Communications Engineering, National Tsing Hua University, Hsinchu 30013, Taiwan \\
\textsuperscript{\rm 2} Department of Electrical Engineering, National Tsing Hua University, Hsinchu 30013, Taiwan \\

\{junteng0211, jerrynthu0602\}@gapp.nthu.edu.tw, chhuang@ee.nthu.edu.tw

}

\usepackage{bibentry}

\begin{document}
	
\maketitle
	
\begin{abstract}
With the improvement of AI chips (e.g., GPU, TPU, and NPU) and the fast development of the Internet of Things (IoT), some robust deep neural networks (DNNs) are usually composed of millions or even hundreds of millions of parameters. Such a large model may not be suitable for directly deploying on low computation and low capacity units (e.g., edge devices). Knowledge distillation (KD) has recently been recognized as a powerful model compression method to decrease the model parameters effectively. The central concept of KD is to extract useful information from the feature maps of a large model (i.e., teacher model) as a reference to successfully train a small model (i.e., student model) in which the model size is much smaller than the teacher one. Although many KD methods have been proposed to utilize the information from the feature maps of intermediate layers in the teacher model, most did not consider the similarity of feature maps between the teacher model and the student model. As a result, it may make the student model learn useless information. Inspired by the attention mechanism, we propose a novel KD method called representative teacher key (RTK) that not only considers the similarity of feature maps but also filters out the useless information to improve the performance of the target student model. In the experiments, we validate our proposed method with several backbone networks (e.g., ResNet and WideResNet) and datasets (e.g., CIFAR10, CIFAR100, SVHN, and CINIC10). The results show that our proposed RTK can effectively improve the classification accuracy of the state-of-the-art attention-based KD method.
\end{abstract}
	
\section{Introduction}\label{sec:introduction}
	Although the deep neural networks (DNNs) have achieved dramatic improvements for many scenarios, most of them contain millions or even billion parameters unsuitable to directly deploy large-scale models on edge devices (e.g., smartphones). The intuitive idea is to design a method to reduce the number of model parameters while maintaining its performance. Recently, many kinds of research, such as pruning weight~\cite{pruning}, quantizing parameters in a fewer bits manner~\cite{quantizing}, and knowledge distillation (KD)~\cite{knowledgedistillation-1,knowledgedistillation-2} have been presented to deal with the issues mentioned above. Among them, KD has been proven to be a promising way to transfer some critical information from a large, well-trained model (i.e., teacher) to a smaller one (i.e., student). The distilled knowledge from the teacher model can be further transformed into soft probabilities, which contain more helpful information than the class label, to guide the student model to achieve better performance than the original scratch.  
	
\begin{figure*}[t]
    \centering
    \includegraphics[width=1\textwidth]{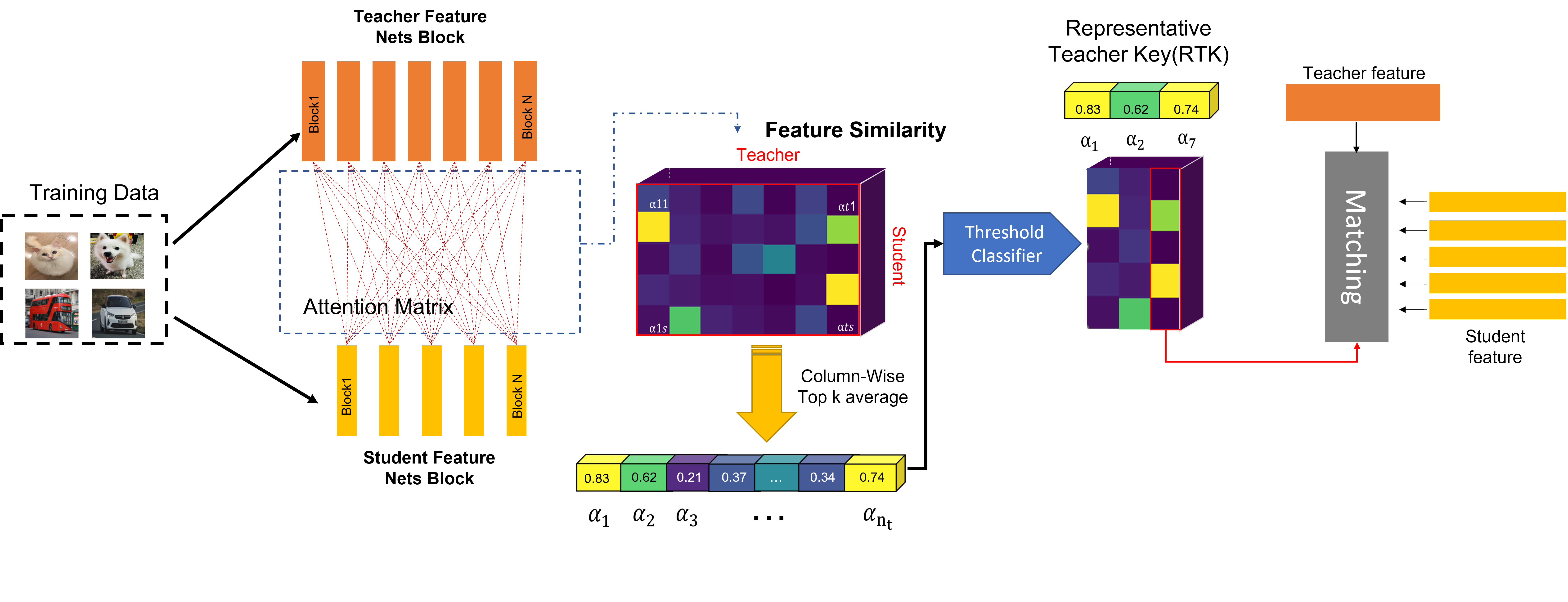}
    \caption{The overall framework of our proposed method. We first generate the similarities of each combination feature pair between the teacher and student models. Then, we calculate the scores representing how much information the teacher feature can contribute to the corresponding student by the column-wise top-$k$ average. Then, the representative teacher key (RTK) and the corresponding representative matrix can be determined by the threshold classifier $\tau_c$. Finally, knowledge from each teacher feature in a representative matrix is transferred to the student.}
\label{fig}
\end{figure*}
	
	Apart from the original form of KD, which transfer output probability distributions of the teacher~\cite{knowledgedistillation-2}, various types of distillation methods, such as using the intermediate-level hints to guide the student~\cite{FitNet}, transferring the attention map from one network to another~\cite{Attentionmap}, and some other approaches~\cite{RKD, CRD} were proposed. However, most proposed methods manually determine what features in the teacher model should be distilled and do not consider the feature similarity between the teacher model and the student model. Therefore, it may provide useless knowledge to mislead the training process of the student model. Also, based on learning to transfer (L2T) research~\cite{L2T}, the result show that KD with the specific pairs can achieve better results than those methods with manually chosen ones. Besides, attention mechanisms~\cite{NeuralImageCaptionGenerationwithVisualAttention} have been considered to improve the model performance in recent years~\cite{attentionreview}. The idea of a sparse attention mechanism has been designed to reduce the memory and computation requirements~\cite{Sparsematrix, Bird}. Also, the low rank-based method~\cite{Lineformer} has been applied to find the important key to reduce the redundant columns. Furthermore, the attention-based feature distillation (AFD)~\cite{AFD} was proposed to identify the importance of links between the teacher model and the student model.
	
	Inspired by the above-mentioned attention mechanism-based works, we come up with a method (see Fig.~\ref{fig}) that is based on the feature similarity and attention mechanism to effectively improve the performance of the work proposed in~\cite{AFD}. We can formulate the similarity values between the corresponding layers in the teacher model and the student model as a simplified attention matrix, allowing the student model to capture more similar knowledge from the teacher model by selecting a specific layer. We also conduct a series of experiments to demonstrate the effectiveness of our proposed method by evaluating extensive experiments on various benchmark datasets (e.g., SVHN~\cite{SVHN}, CIFAR ~\cite{CIFAR}, and CINIC~\cite{CINIC}) and different network architectures. The experimental results show that our proposed method outperforms the state-of-the-art attention-based KD method. We summarize our contributions as follows:
	
	\begin{itemize}
		\item  We propose a novel method, namely RTK, to further improve the performance of the AFD method. The main idea of our RTK is to link the concept of attention mechanism and the top-$K$ value to select the representative feature keys with a threshold classifier. 
		\item We embed a down-sampling block into our network to effectively preserve the original teacher features for the procedure of KD.
		\item  We verify the effectiveness of our RTK method by conducting a series of experiments with several image classification datasets and different network architectures. The results show that the RTK method can well preserve the information from the teacher model to train the student model with a few amount of performance loss.
	\end{itemize}
	
	The rest of this paper is organized as follows. Section~\ref{sec:relatework} introduces the related works. Section~\ref{sec:method} introduces our proposed method. Section~\ref{sec:experimentalresult} shows the experimental results of our proposed method on several benchmark datasets. Final conclusion will be drawn in Section~\ref{sec:conclusion}.
	
	\begin{figure*}[t]
		\centering
		\includegraphics[width=1\textwidth]{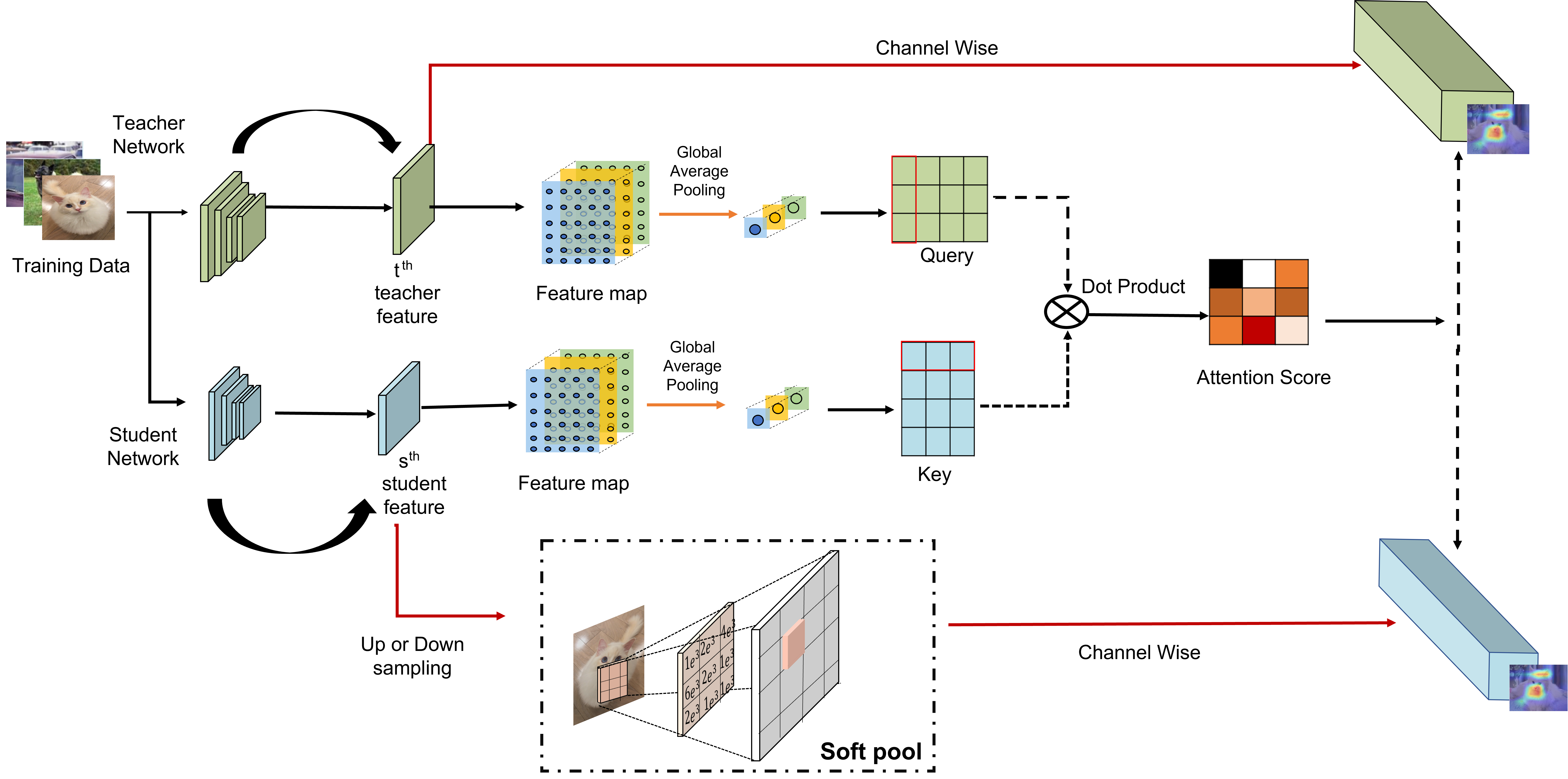}
		\caption{Overview of the proposed network. The feature, obtained by globally pooled, can be used to estimate the similarities, which can help us to identify the specific representative teacher key. Then, we add the up-sampling/down-sampling method to adjust the dimension of the selected feature pairs. Finally, we use the channel-wise strategy to calculate the distance between the features.}
		\label{fig:proposedframwork}
	\end{figure*}
	
	\section{Related Works}\label{sec:relatework}
	\subsection{Attention Mechanism}
	Instead of considering the entire features from input data, the core idea of attention mechanisms is to focus on the components located in some specific regions. Recently, the first attention mechanism-based machine translation method was proposed in~\cite{attentionmechanism_machinetranslate}. The attention mechanism is applied to learn the model to conduct align and translate tasks jointly. In~\cite{NeuralImageCaptionGenerationwithVisualAttention}, the visual attention mechanism is utilized to automatically discover the crucial features to generate the captions of images. A novel co-attention model for visual question answering that jointly interprets image and question attention was proposed in~\cite{videoquestion}. The attention mechanisms can improve not only model performance but also a strategy to make the process of model operation explainable~\cite{NeuralImageCaptionGenerationwithVisualAttention,attentionisallyouneed}. 
	
	\subsection{Knowledge Distillation}
	Knowledge distillation (KD) was first introduced to compress the DNNs by transferring the dark knowledge (i.e., the distribution of logits) from the teacher model to the student model~\cite{knowledgedistillation-1,knowledgedistillation-2}. With this dark knowledge, the student model may achieve comparable performance with the teacher model even though we significantly reduce the number of model parameters. In recent years, a new metric of intermediate feature maps between the teacher model and the student model was proposed in~\cite{FitNet}, and a regressor was also presented to match the different dimensions between the teacher model and the student model. Besides, the spatial attention map extracted from the intermediate layers based on the gradient and the activation information was proposed in~\cite{Attentionmap}. The flow of solution procedure (FSP) matrix calculated by Gram matrix of feature maps from the corresponding layers was designed to transfer knowledge information in~\cite{FSP} well. However, the methods mentioned above ignored the relationship among the corresponding instances, which may be positive information to be considered to improve the student model accuracy. Hence, a KD method using the relation between data instances to guide the learning process of the student model was proposed in~\cite{RKD, CorrelationCongruence, SP}. Another technique is discussed about relational KD with contrastive learning~\cite{CRD}. In~\cite{VID, Likewhatyoulike} using different regularizations to transfer knowledge to the student model. However, as most of the prior works are manually linked the intermediate features	between the teacher model and the student model, research in~\cite{AFD} provides an innovative idea to deal with it. In our work, we offer a different perspective based on the strategy used in~\cite{AFD} and conduct a series of experiments to demonstrate our method can achieve better performance.

	\section{Proposed Method}\label{sec:method}
	In this section, we first introduce the problem formulation of attention-based KD. Then we will illustrate the main concept of our proposed method to explain how we improve the method proposed in~\cite{AFD}. Finally, the overall operation procedure of our proposed method is presented in Fig.~\ref{fig:proposedframwork}. We denote $\mathfrak{M}_{\mathrm{{T}}}$ = $\{m_{{1}}^\mathrm{{T}} ,...,m_{{n_t}}^\mathrm{{T}}\}$, $m_i^{\mathrm{{T}}} \in \mathbb{R}^{h^{\mathrm{{T}}}\times w^{\mathrm{{T}}}\times c^{\mathrm{{T}}}}$ as a set of feature maps from teacher model and $\mathfrak{M}_{\mathrm{{S}}}$ = $\{m_{{1}}^\mathrm{{S}} ,...,m_{{n_s}}^\mathrm{{S}}\}$, $m_j^{\mathrm{{S}}} \in \mathbb{R}^{h^{\mathrm{{S}}}\times w^{\mathrm{{S}}}\times c^{\mathrm{{S}}}}$ as a set of feature maps from student model where $n_t$, $n_s$, $h$, $w$, and $c$ represent the number of the feature maps from teacher model, the number of the feature maps from student model, height, width, and channel dimension, respectively. 
	The total number of combination pairs of feature map is ($n_t$ $\times$ $n_s$). In~\cite{AFD}, they took all of the combination pairs into account for KD, which may include the irrelevant feature maps that could lower the quality of distillation result. Instead, we propose a novel method to select the combination pairs that are more relevant between the teacher model and the student model. 
	
	First, we use the similar method applied in~\cite{AFD} to determine the attention values of each combination pair ($m_{{i}}^\mathrm{{T}}$, $m_{{j}}^\mathrm{{S}}$). We generate the query vectors $\textbf{q}_s$  and the key vectors $\textbf{k}_s$ by:
	
	\begin{align}
	\label{Query}
	~~\textbf{q}_{{i}}&={\rm relu}(W_i^\mathrm{{Q}}\cdot \phi(m_{{i}}^\mathrm{{T}})),~~i=1,2,\ldots,n_t\\
	\label{Key}
	~~\textbf{k}_{{j}}&={\rm relu}(W_j^\mathrm{{K}}\cdot \phi(m_{{j}}^\mathrm{{S}})),~~j=1,2,\ldots,n_s   
	\end{align}

	where $\phi(.)$ is a global average pooling function, $W_i^\mathrm{{Q}} \in \mathbb{R}^{c\times c_i^{\mathrm{{T}}}}$ is a transition matrix for transiting the $i$-th feature in the corresponding layer of teacher model, and $W_j^\mathrm{{K}} \in \mathbb{R}^{c\times c_j^{\mathrm{{S}}}}$ is a transition matrix for transiting the $j$-th feature in the corresponding layer of student model. Then, the column vectors $\mathbf{\alpha}_i \in \mathbb{R}^{n_s}$ in attention matrix $\bf{A}$ $\in$ $\mathbb{R}^{n_s \times n_t}$ can be determined by:
	
	\begin{align}\label{attentionscore_z}
	z_{i,j} &=  (\textbf{q}_{{i}}^\top W_{{j}}\textbf{k}_{{i,j}}+(\textbf{p}_{{i}}^{{T}})^\top  \textbf{p}_{{j}}^{{S}})/\sqrt{d}\\
	\label{attentionscore_a}
	\mathbf{\alpha}_i &= \left(
	\begin{array}{c}
	\frac{e^{z_{i,1}}}{\mathlarger{\mathlarger{{\scriptstyle\sum}}}_{j=1}^{n_s} e^{z_{i,j}}}\\
	\frac{e^{z_{i,2}}}{\mathlarger{\mathlarger{{\scriptstyle\sum}}}_{j=1}^{n_s} e^{z_{i,j}}}\\
	\ldots \\
	\frac{e^{z_{i,n_s}}}{\mathlarger{\mathlarger{{\scriptstyle\sum}}}_{j=1}^{n_s} e^{z_{i,j}}}
	\end{array}
	\right) \in \mathbb{R}^{n_s}
	\end{align}
	
	where $W_{{j}} \in \mathbb{R}^{c \times c}$ is a bi-linear weight matrix used to handle the attention value from different level of layer, $\textbf{p}_{{i}}^{{T}} \in \mathbb{R}^c$ and $\textbf{p}_{{j}}^{{S}} \in \mathbb{R}^c$ are two positional encoding vectors utilized to assign a unique representation on each position.
	
	Next, we apply the concept of top-$k$ value on each column vector $\mathbf{a}_i$ in the attention matrix $\bf {A}$ we have built. Then, we can determine the impact value by averaging the top-$k$ values in each attention column vector. By considering the impact values, we can know which teacher features are more suitable for conducting the KD process. Here, we use a threshold classifier $\tau_c$ to determine which teacher features should be reserved and utilized in the KD process. We denote these teacher features as representative teacher keys (RTKs) in this paper. 
	
	Finally, our distillation loss $\mathcal{L}_{RTK}$ will transfer the knowledge of RTKs (e.g., $m_{RTK}^T$). Only part of the feature maps in the teacher model will be used to make a contribution to the training process of the student model. The loss function is defined as:
	
	\begin{equation}
	\label{student_loss}
	\mathcal{L}_{RTK}=\Sigma_{{R}}\Sigma_{{s}}\alpha_{{R,s}}\Vert \tilde{\phi^C}(m_{{RTK}}^T)-
	\tilde{\phi^C}(\hat{m}_{{s}}^S)
	\Vert_{{2}}
	\end{equation} 
	
	where $\tilde{\phi^C}$ is a combined function of a channel-wise average pooling layer with L2 normalization~\cite{Attentionmap}. Besides, $\hat{m_{{s}}^S}$ is the up-sampled/down-sampled result from $m_{{s}}^S$ to match the feature map size of the RTKs. Here, we apply a memory-efficient pooling method Softpool~\cite{Softpool} as our up-sampling/down-sampling method for the distillation process. Therefore, the up-sampled/down-sampled feature map $\hat{m_{{s}}^S}$ can be generated by:
	\begin{equation}
	\label{softpool}
	\begin{gathered}
	\hat{m_{{s}}^S}\tilde{(a)}=m_{{s}}^S\left(\sum_{{i\in R}}\frac{(e^{a_{i}} \times a_{{i}})}{\sum_{j\in R}e^{a_{j}}}\right)
	\end{gathered}
	\end{equation}
	where $a$ is the activation map of student feature $\hat{m_{{s}}^S}$.
	
	Finally, our total loss function $\mathcal{L}_{TL}$ includes three elements as:
	\begin{equation}
	\label{total_loss}
	\begin{gathered}
	\mathcal{L}_{TL} = \mathcal{L}_{CE} + \alpha \mathcal{L}_{KL}(p^s(T),p^t(T)) + \beta \mathcal{L}_{RTK}
	\end{gathered}
	\end{equation} 
	where $p^t$ and $p^s$ are the softened targets of teacher and student network with temperature parameter $T$, respectively. Here, we use Kullback-Leibler (KL) divergence loss to measure them since the softened results can be viewed as probability distributions. Besides, we use cross-entropy loss $\mathcal{L}_{CE}$ to measure the classification loss. The $\alpha$ and $\beta$ are two hyperparameters used to control which distillation loss term is more influential during the training procedure. The overall training procedure with our proposed RTKs is illustrated in Algorithm~\ref{alg:alogorithm1}.
	
	\begin{algorithm}[tb]
		\caption{The framework of model training with our proposed RTK method}
		\label{alg:alogorithm1}
		\textbf{Input}: Teacher model $\mathcal{M}_t$, student model $\mathcal{M}_s$ and dataset $\mathcal{D}=[(X_{{i}},y_{{i}})_{{i=1}}^N]$\\
		\textbf{Parameter}: $\Theta_{pretrain}$, $\Theta_T$, $\Theta_S$\\
		\textbf{Hyperparameters}: $\alpha$,~$\beta$,~$T$,~$m$\\
		\textbf{Output}: $\mathcal{M}_s$ with $\Theta_S$
		\begin{algorithmic}[1] 
			\STATE \textbf{Stage 1:} Training $\mathcal{M}_t$
			\STATE $\Theta_{pretrain}$ $\gets$ random initialization
			\STATE \textbf{Repeat}
			\STATE $\mathcal{L} \gets $ $\mathcal{L}_{CE}(\mathcal{M}_t(X_i), y_{i})$\;
			\STATE Update $\Theta_{pretrain}$ by considering $\mathcal{L}$\;
			\STATE \textbf{Until} Finish the training process of the teacher model
			
			\STATE \textbf{Stage 2:} Training $\mathcal{M}_s$ with RTK method
			\STATE $\Theta_T$ $\gets$ $\Theta_{pretrain}$ from \textbf{Stage 1}
			\STATE $\Theta_S$ $\gets$ random initialization
			\FOR{ $i \gets 1$ to $epoch$}
			\STATE Compute $\bf {A} \gets$ Equation (\ref{Query})-(\ref{attentionscore_a})\;
			\FOR{ $j \gets 1 $ to $n_s$ }
			\STATE Compute column-wise top-$K$ average value $\overline{s_j}$ for the corresponding (\textbf{q}, \textbf{k}) pair\;
			\IF{$\overline{s_j}> \tau$}
			\STATE keep the j-th feature as the RTK to distill\;
			\ELSE
			\STATE continue\;
			\ENDIF
			\ENDFOR
			\STATE Update $\bf {\tilde{A}}$ with RTKs\;
			\STATE $\mathcal{L}_{TL}$ $\gets$ Equation (\ref{total_loss})\;
			\STATE Update $\Theta_S$ by considering $\mathcal{L}_{TL}$\;
			\ENDFOR
			\STATE \textbf{return} $\mathcal{M}_s$ with $\Theta_S$
		\end{algorithmic}
	\end{algorithm}
	
	\begin{table*}[ht]
		\centering
		\begin{tabular}{|c| c| c| c|c|} 
			\hline
			Method & Model & Params & CIFAR10($\%$) & CIFAR100($\%$) \\ [0.5ex] 
			\hline\hline
			Student & ResNet-20 & 0.27M&92.37 & 69.89 \\
			
			KD~\cite{knowledgedistillation-2} & ResNet-20 & 0.27M&93.28 & 70.98 \\
			
			FitNet~\cite{FitNet} & ResNet-20 & 0.27M&92.56 & 70.05\\
			
			AT~\cite{Attentionmap} & ResNet-20 & 0.27M&93.00 & 70.56 \\
			
			RKD~\cite{RKD} & ResNet-20 & 0.27M&93.29 & 70.43 \\
			
			CRD~\cite{CRD} & ResNet-20 & 0.27M&92.66 & 70.95\\
			
			AFD~\cite{AFD} & ResNet-20 & 0.27M & 93.26 & 71.53\\
			
			Ours & ResNet-20 & 0.27M&\textbf{93.40} & \textbf{71.75} \\
			\hline\hline
			Teacher & ResNet-56 & 0.85M&94.08 & 72.81 \\
			\hline
		\end{tabular}
		
		\caption{Experiment on CIFAR10 and CIFAR100 datasets with different knowledge distillation methods. We conduct the Experiments using the ResNet-56 as Teacher Network and the ResNet-20 as Student Network.} \label{Table1}
	\end{table*} 
	
	\section{Experimental Result}\label{sec:experimentalresult}
	In this paper, we select a couple of datasets to demonstrate the effectiveness of our proposed method. Besides, we choose the benchmark datasets, including SVHN~\cite{SVHN}, CIFAR10, CIFAR100~\cite{CIFAR}, and CINIC10~\cite{CINIC} as the experimental datasets. In all experiments, we apply data augmentation methods (e.g., random crop and padding horizontal flipping) to extend the data quantity.
	\subsection{Datasets}
	\textbf{SVHN}~\cite{SVHN}: The Street View House Numbers (SVHN) dataset is a real-world image dataset widely used in object detection algorithms with minimal data preprocessing requirements. The SVHN contains ten categories RGB 32×32 digit images from ‘0’ to ‘9’, and there are 73,257 digits for training and 26,032 digits for testing.
	
	\noindent\textbf{CIFAR}~\cite{CIFAR}: The CIFAR is an object detection dataset composed of objects common in reality. There are 60,000 32×32 RGB images in the CIFAR dataset (50,000 images for training and 10,000 images for testing). The CIFAR dataset can be divided into CIFAR10 and CIFAR100 depending on the number of categories of objects. CIFAR10 contains ten types of objects, and each class has 6,000 pictures. CIFAR100 includes 100 categories of objects, and each category has 600 pictures.
	
	\noindent\textbf{CINIC10}~\cite{CINIC}: The CINIC10 dataset consists of images from both the CIFAR dataset and the ImageNet dataset. The dataset contains 270,000 images at a spatial resolution of 32 × 32 via the addition of downsampled ImageNet images. We adopt the CINIC10 dataset for the experimentation because its scale is closer to the case of ImageNet, and it is a noisy dataset.
	
  \begin{figure}[htbp]
		\centering
		\subfigure[The result of Cifar10.]{
			\includegraphics[width=1\linewidth]{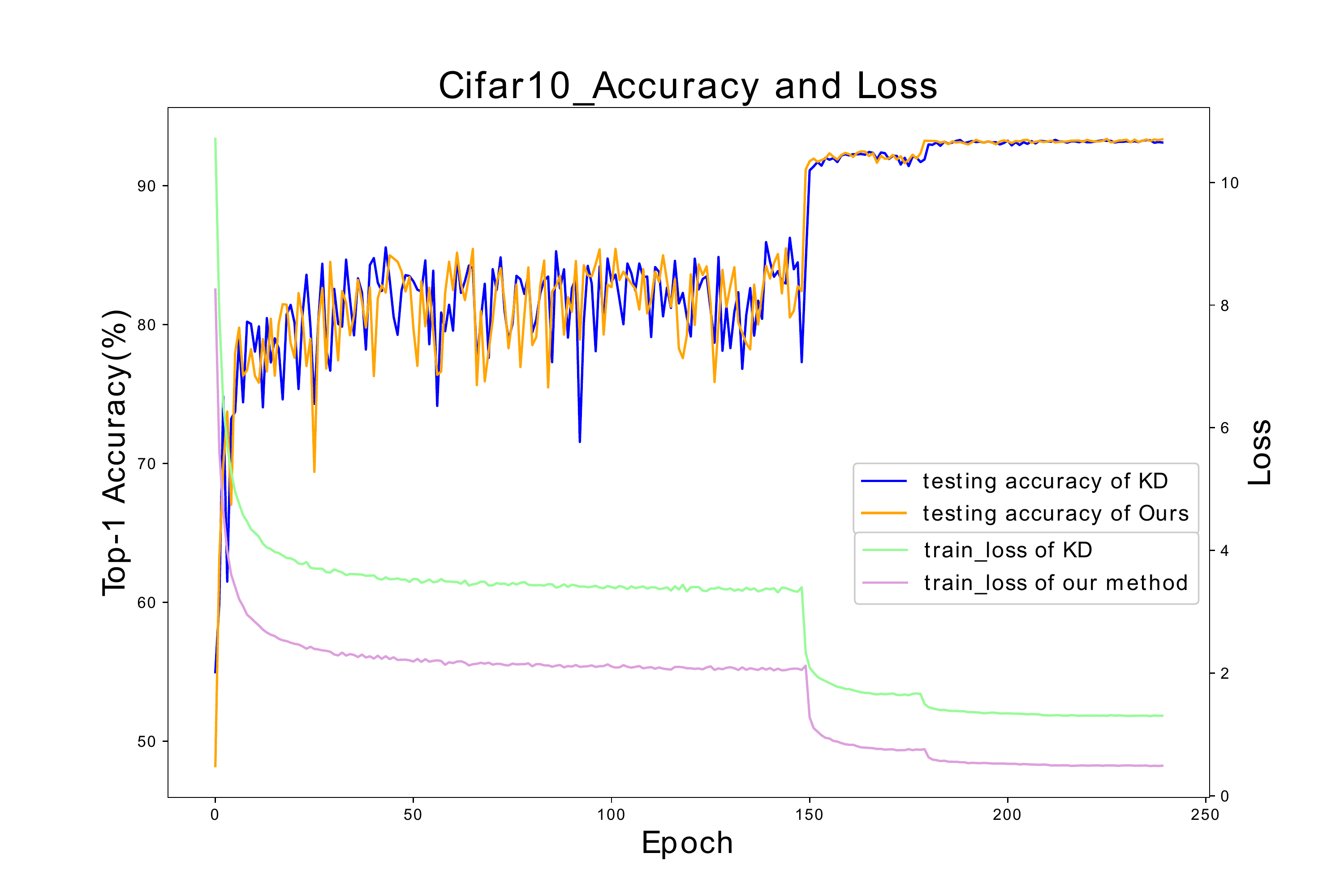}
		}
		\subfigure[The result of Cifar100.]{
			\includegraphics[width=1\linewidth]{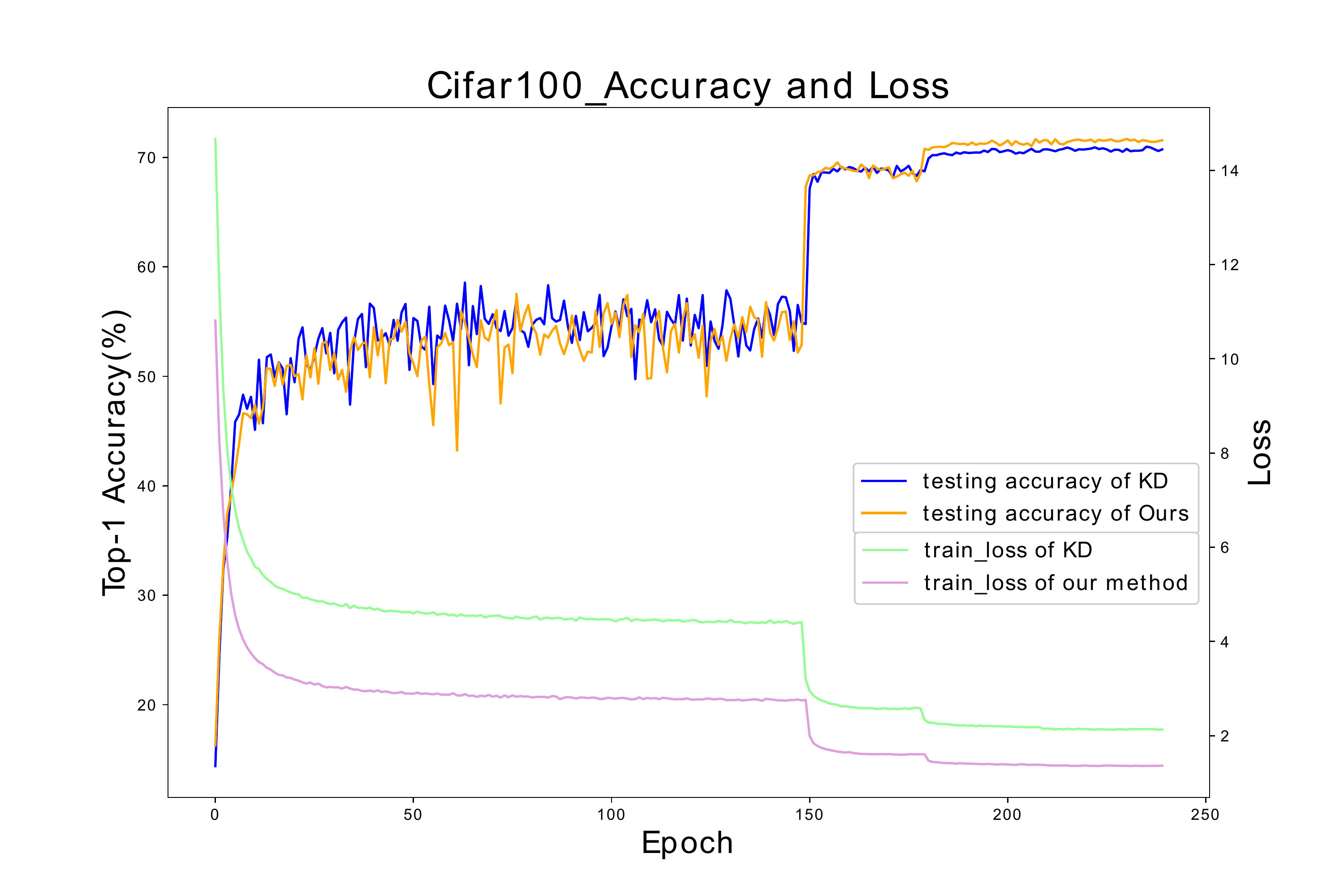}
		}
		\caption{The training loss and testing accuracy of conventional KD and our method on the CIFAR10/CIFAR100 dataset. In this experiment, we adopt Resnet-56 as the teacher network and Resnet-20 as the student network. Both of them are trained with epoch 240. (a) The result of the experiment on the CIFAR10 Dataset (b) The result of the experiment on the CIFAR100 Dataset.}
		\label{Figure3}
	\end{figure}

\begin{figure*}[htbp]
    \centering
    \includegraphics[width=1\linewidth]{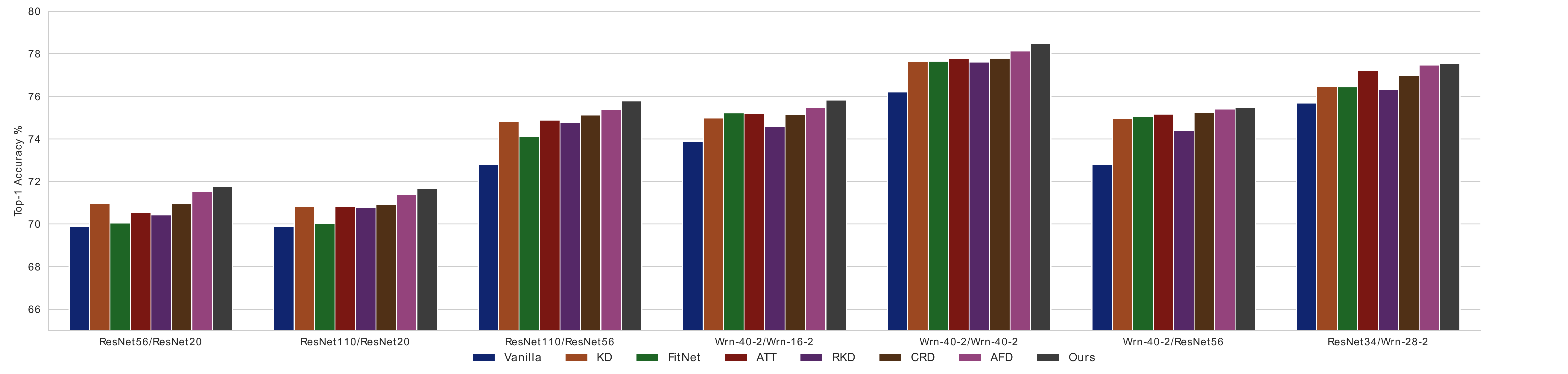}
    \caption{Knowledge distillation with different network structure pairs (teacher/student) on Cifar100. The result shows that our method can train the student model in a good shape regardless of the architecture. }
    \label{Figure4}
\end{figure*}

    \begin{figure}[htbp]
		\centering
		\includegraphics[width=1\linewidth]{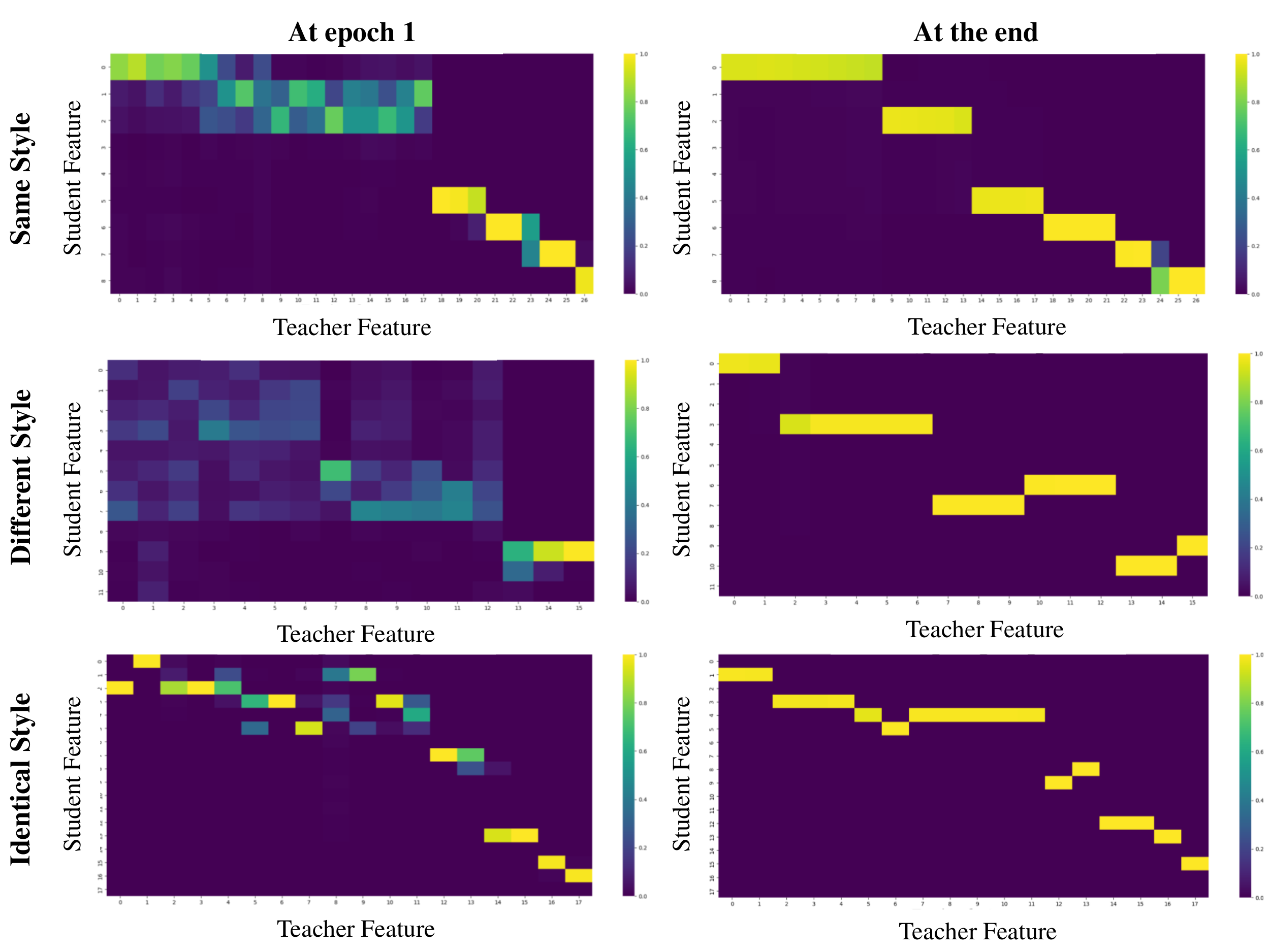}
		\caption{Attention matrices for our proposed method. Each matrix is the average overall $z_{i,j}$ at one of the training epochs. Rows and columns of matrices represent the student and teacher features, respectively. The results are three compared distillation settings of our method on the Cifar100 dataset.; ResNet-56$\rightarrow$ ResNet-20 \emph{Same Architecture}, ResNet-34$\rightarrow$ WRN-28$\times$2 \emph{Different Architecture}, and WRN-40$\times$2 $\rightarrow$ WRN-40$\times$2 \emph{Self Architecture}. }
		\label{Figure6}
	\end{figure}

	\begin{table*}[htbp]
		\renewcommand\arraystretch{1}
		\centering
		\scalebox{1}{
			\begin{tabular}{|c|c|c|c|c|c|} 
				\hline
				Model & Method & CIFAR10 & CIFAR100 & SVHN&CINIC10 \\ [0.5ex] 
				\hline\hline
				Teacher & ResNet-56 & 94.08&72.81 & 95.05&83.56 \\
				
				Student(ORI) & ResNet-20 & 92.37&69.89 & 92.77&81.80 \\
				
				Student(Ours) & ResNet-20 & \textbf{93.40}&\textbf{71.75} & \textbf{94.36}&\textbf{83.35}\\
				\hline
				Teacher & ResNet34 & 95.45&79.43 & 98.56&86.71 \\
				
				Student(ORI) & WRN$\times$28$\times$2 & 94.78&75.69 & 95.50&85.19 \\
				
				Student(Ours) & WRN$\times$28$\times$2 & \textbf{95.35}&\textbf{77.56} & \textbf{96.43}&\textbf{85.95}\\
				\hline
				
				Teacher & WRN$\times$40$\times$2 & 94.84&76.20 & 96.15&85.33 \\
				
				Student(ORI) & WRN$\times$40$\times$2 & 94.84&76.20 & 96.15&85.33 \\
				
				Student(Ours) & WRN$\times$40$\times$2 & \textbf{95.39}&\textbf{78.47} & \textbf{96.49}&\textbf{86.57}\\
				\hline
			\end{tabular}
		}
		\caption{The Experiment results of the student network trained by our method on the benchmark dataset. The Student ORI denotes the result of the original student network.} \label{Table2}
	\end{table*}

	\begin{table}[htbp]
		\renewcommand\arraystretch{1}
		
		\centering
		\scalebox{0.90}{
			\begin{tabular}{|c|c|c|c|c|} 
				\hline
				Method & CIFAR10 & CIFAR100 & SVHN&CINIC10 \\ [0.5ex] 
				\hline\hline
				Our method & \textbf{0.959}&\textbf{0.856} & \textbf{0.970}&\textbf{0.906}
				\\
				\hline
				Teacher & 0.962 & 0.864 & 0.973 & 0.907\\
				
				Student & 0.953 & 0.844 & 0.963 & 0.899\\
				\hline
			\end{tabular}
		}
		
		\caption{Average ROC Performance Metrics of Teacher and Student model on benchmark datasets. The teacher network and the student network are ResNet56 and ResNet20, respectively.} \label{Table3}
	\end{table}

	\subsection{Experiment Setup}
	\textbf{Network architecture}: In our CIFAR100 experiment, we select the famous ResNet-Family networks and WideResNet-Family networks as our backbone model to train on the above-mentioned datasets with several different cases. We test different combination pairs (teacher, student) including (ResNet-56, ResNet-20), (ResNet-110, ResNet-20), (ResNet-110, ResNet-56), (WRN-40$\times$2, WRN-16$\times$2), (WRN-40$\times$2, WRN-40$\times$2), (WRN-40$\times$2, ResNet-56), and (ResNet-34, WRN-28$\times$2) seven cases. In other dataset experiments, we provide three representative cases (e.g., combination pairs) to demonstrate our proposed method. 
	
	\noindent\textbf{Implementation details}: For all of the experiments on CIFAR10, CIFAR100, and CINIC10, we set the mini-batch size as 64 and the optimizer as stochastic gradient descent (SGD) with hyperparameter momentum $m=0.9$ to train the model with 240 epochs. The learning rate and weight decay are set as 0.01 and 0.05 and divided by ten at the 150-th, 180-th, and 210-th epoch. For the cases on the SVHN dataset, we change the batch size to 128, the maximum iteration to 50 epochs, and the learning rate is set as 0.001. For the results of baseline methods, we reproduce them by using the provided official code and the same environment setting of their experiments. 
	
	\subsection{Evaluation of our method}
	\textbf{Results on CIFAR}: We set the ResNet-56 pre-trained model which classification accuracy is  $94.08\%$ and $72.81\%$ as the teacher model. Then, we apply our proposed method and other existing methods to distill the knowledge from it to train the student models. Table~\ref{Table1} shows the results of the experiment on CIFAR10 and CIFAR100. We know that our method brings effective knowledge to improve the performance of the student model with $1.03\%$ and $1.86\%$ for CIFAR10 and CIFAR100, respectively. Besides, we successfully reduce the number of the model parameter from 0.85M to 0.27M, which is a 3X compression rate, with only $0.68\%$ performance loss. Fig.~\ref{Figure3} show the training loss and testing accuracy of the conventional KD and our method. Here, we train the model with the same setting for comparison. It is worth noting that since our proposed method is to select the feature candidate between the teacher model and the student model, it may have different results if we consider other network structures. 
 
    Moreover, to demonstrate the generality of our method, we evaluate our proposed method with a variety of network structures, including ResNet~\cite{ResNet} and its variants WideResNet~\cite{WideResNet} on CIFAR100. The results are presented in Fig.~\ref{Figure4}. From the results of the experiment, our method outperforms the existing techniques on CIFAR10 and CIFAR100.
	
	\begin{figure*}[t]
		\centering
		\includegraphics[width=0.8\linewidth]{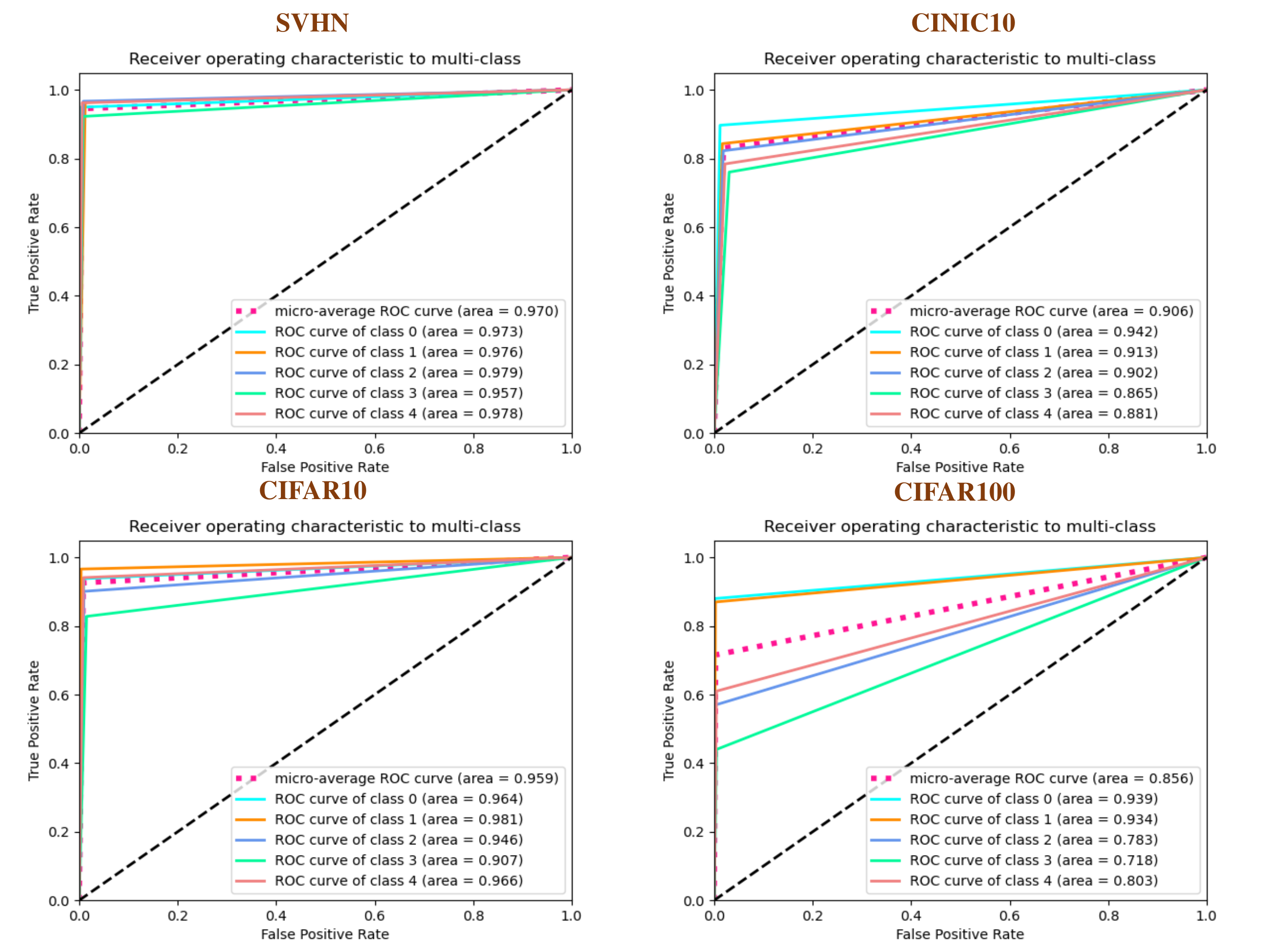}
		\caption{Multiclass ROC curves with deep neural networks applied of our method on four datasets: Cifar100, Cifar10, SVHN, and CINIC10. The results are based on one of our distillation settings: ResNet-56$\rightarrow$ ResNet-20. The micro-average is derived from the ROC curve. Besides, we select five classes for each dataset to show their AUC scores. For CIFAR10 as an example, Class 0: plane, Class 1: car, Class 2: bird, Class 3: cat, and Class 4: deer.}
		\label{Figure7}
	\end{figure*}	
	
	\noindent\textbf{Results on SVHN and CINIC10}: In this part, we use a similar setting as the experiment of CIFAR. Specifically, we further conduct experiments on different network structures on SVHN and CINIC10 to verify the effectiveness of our method. We classify the experiments into three scenarios. The first one, called \emph{same architecture style}, is to conduct the KD process with the same network structure, and the second one, called \emph{different architecture style}, runs the KD process with a different network structure. The last one, called \emph{identical architecture style}, is to distill the knowledge from the same network structure and the same model size to improve the classification performance. Table~\ref{Table2} shows the results of the aforementioned cases, and our method can improve the classification accuracy even though the network structure between teacher and student is different.
	
	\noindent\textbf{Visualization and analysis}:
	Here, we provide the visualization of the attention matrix to illustrate the experiment results, see Fig.~\ref{Figure6}. To observe the variability of attention matrices $\bf {A}$, we analyze the aforementioned three architecture pairs by conducting the same experiment in \cite{AFD}. Fig.~\ref{Figure6} shows the updating pattern of $\bf {A}$ during the training phase. Although the similarity will change in each training run and the feature candidate may be different, our RTK method can mitigate such a situation to effectively find the more important feature to distill. 
 Finally, we evaluate the average ROC-AUC value on the four datasets with the pair (ResNet-56, ResNet-20) as an example. The model is well trained if the ROC curve is near the upper-right point. In Fig.~\ref{Figure7}, we present the ROC curves of our method on the four datasets. The results show that our RTK-based distillation model can reach a performance relatively close to the performance of the teacher model. Table~\ref{Table3} shows all the results of ROC performance metrics on benchmark datasets.
	
	\section{Conclusion}\label{sec:conclusion}
	In this paper, we propose a novel representative teacher key (RTK) strategy based on an attention mechanism to effectively conduct the knowledge distillation (KD) for the training process of the student model. Besides, we apply our RTK to various network architectures to demonstrate its effectiveness and generality. The experimental results show that the student model, trained with RTKs, can achieve performance gains in image classification tasks. In recent years, decentralized learning has become a key research topic due to the importance of data privacy. Therefore, in future works, we aim to combine our proposed KD method with federated learning (FL) and blockchain to design a privacy-preserving distributed learning system. 

	
\bibliography{aaai23}
\end{document}